\documentclass[journal]{IEEEtran}

\usepackage{float}
\usepackage{graphicx,subfigure} 
\usepackage{amsmath}
\usepackage{amsthm}
\usepackage{amssymb}
\usepackage{gensymb}
\usepackage{enumerate}
\usepackage{csquotes}
\usepackage{url}
\usepackage{blox}
\usepackage{breakurl}
\usepackage{blindtext}
\usepackage{rotating}
\usepackage{color}

\begin{document}
\title{Decentralized Control Systems Laboratory Using Human Centered Robotic Actuators}
% \title{Remote Control Systems Laboratory Using Human-Centered Robotic Actuators}

% \title{A Remote STEM Education Architecture Using State-of-the-art
% Laboratory Hardware}

% \author[1*]   {Binghan He}
% \author[1]   {Kunye Chen}
% \author[1]   {Rachel Schlossman}
% \author[2]   {Neal Ormsbee}
% \author[2]   {Mara Altman}
% \author 	 {Nathan Young}
% \author[2]   {Matt Mangum}
% \author[1]   {Luis Sentis}
% \affil [1]   {Authors are with The Departments of Mechanical Engineering (B.H., K.C., R.S., N.Y.) or Aerospace Engineering (L.S.), University of Texas at Austin, Austin, TX 78712-0292, USA}
% \affil [2]   {Authors are with The Faculty Technology Studio, University of Texas at Austin, Austin, TX 78712-0292, USA}

% \affil [*]   {binghan@utexas.edu}

% \affil[+]{these authors contributed equally to this work}

%\keywords{Keyword1, Keyword2, Keyword3}

%\begin{document}
\author{Binghan~He$^1$, Kunye~Chen$^1$, Rachel~Schlossman$^1$, Neal~Ormsbee$^2$, Mara~Altman$^2$,\\ Nathan~Young$^1$, Matt~Mangum$^2$ and~Luis~Sentis$^3$% <-this % stops a space
\thanks{$^{1}$ Authors are with The Department of Mechanical Engineering, University of Texas at Austin, Austin, TX 78712, USA. Send correspondence to {\tt\small binghan@utexas.edu}.}% <-this % stops a space
\thanks{$^{2}$ Authors are with The Faculty Technology Studio, University of Texas at Austin, Austin, TX 78712, USA.}
\thanks{$^{3}$ Author is with The Department of Aerospace Engineering and Engineering Mechanics, University of Texas at Austin, Austin, TX 78712, USA.}% <-this % stops a space
}

\flushbottom
\maketitle
%\thispagestyle{empty}
\begin{abstract}
University laboratories deliver unique hands-on experimentation for STEM students but often lack state-of-the-art equipment and provide limited access to their equipment. The University of Texas Cloud Laboratory provides remote access to a cutting-edge series elastic actuators for student experimentation regarding human-centered robotics, dynamical systems, and controls. Through a browser-based interface, students are provided with various learning materials using the remote hardware-in-the-loop system for effective experiment-based education. This paper discusses the methods used to connect remote hardware to mobile browsers, the adaptation of textbook materials regarding system identification and feedback control, data processing to generate clean and useful results for student interpretation, and initial usage of the end-to-end system for individual and group learning. 
\end{abstract}
\section{Introduction}

In control system and robotic laboratory courses, students often grasp novel concepts and ideas from experimental data and analysis of experiments. The strategies for conducting these laboratories, however, have not changed in decades. First, advanced state-of-the-art devices can be prohibitively expensive and require significant human supervision. While there are some exceptions \cite{mondada2009puck}\cite{zhang2008development}\cite{bischoff2011kuka}, there will always exist a segment of the population for which even the relatively low-priced systems are out of reach. In parallel, many top-performing devices and testbeds developed from competitive research and industrial projects are rarely used for educational purposes but hold a strong potential for such purpose. It is compelled to enhance control systems education by creating methods for accessing ubiquitously cutting-edge equipment.

% Laboratories are typically organized into lab sections, which imply time and location limitations. However, not everyone who strives to learn STEM subjects has the availability and the means to register in research-oriented schools. 

% In STEM courses and high-tech training courses, students often grasp novel concepts and ideas from experimental data and analysis of laboratory experiments. The strategies for conducting these laboratories, however, have not changed in decades and are not keeping up with increased demand, especially in distance learning settings. First, advanced laboratory devices can be prohibitively expensive and require significant human supervision. While there are some exceptions, there will always exist a segment of the population for which even the relatively low-priced systems are out of reach. In parallel, many top-performing devices and testbeds developed from competitive research projects stop being used after the main discovery has been made. Those devices could continue being used for educational purposes past their research lifetime. It would be difficult to fulfill the demands of future STEM learners without sharing laboratory resources. Laboratories are typically organized into lab sections, which imply time and location limitations. However, not everyone who strives to learn STEM subjects has the availability and the means to register in research-oriented schools. 

\begin{figure}[t]
\centering
\includegraphics[width=\linewidth]{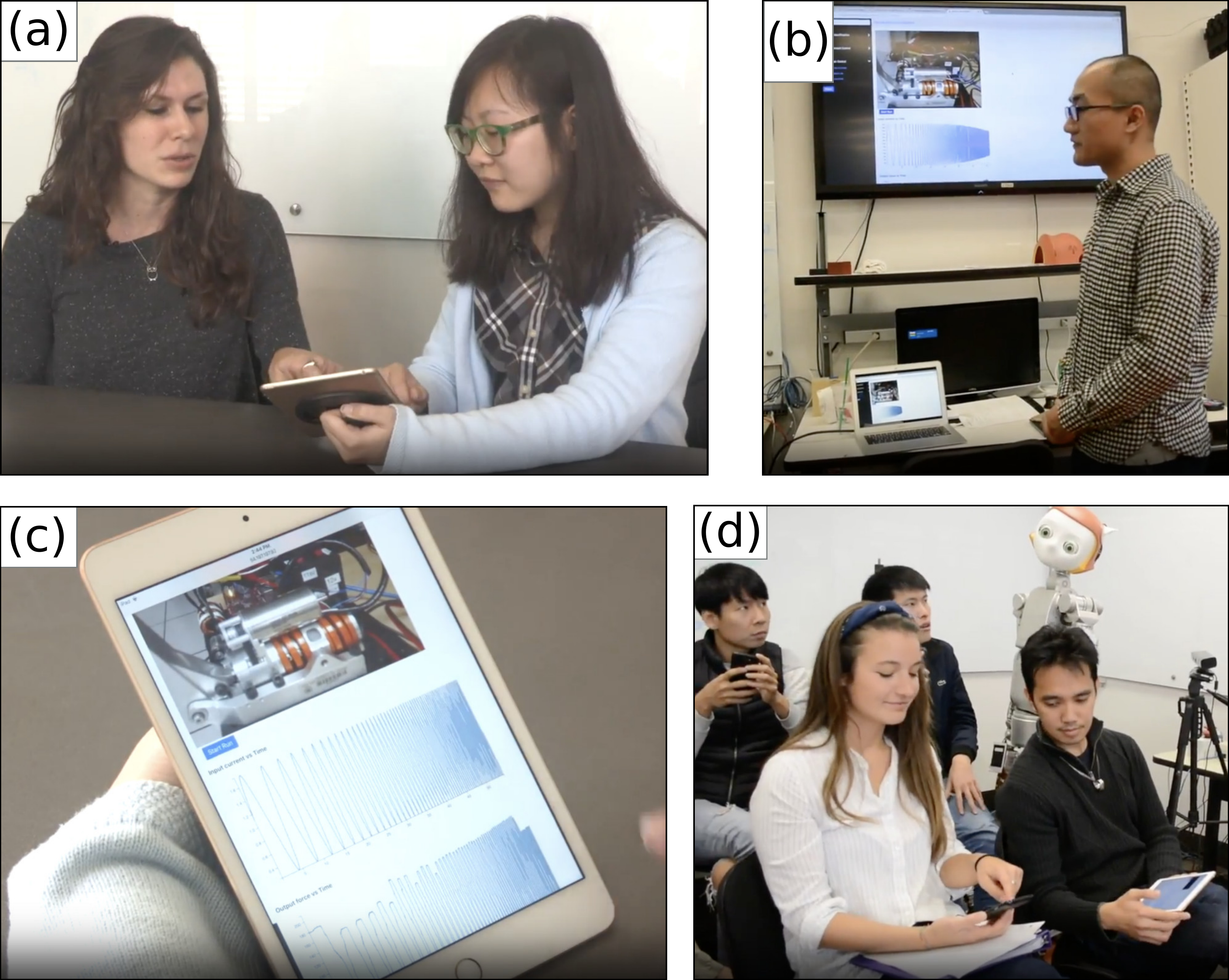}
\caption{(a) A demonstration of usage of the CLAB system, in which a teaching assistant and a student perform remote experiments and discuss the results. (b) Another teaching assistant delivers a lecture using the CLAB remote laboratory. (c) A snapshot of the CLAB interface accessed from an iPad, with live video streaming and live data provided below. (d) Multiple students remotely accessing the live experiments from their respective mobile devices during class time.}
\label{fig:overview}
\end{figure}

In the past, several projects have been geared towards an online laboratory for education. An Internet based control laboratory is devised in \cite{hahn2000remote}, consisting of eighteen benches containing educational equipment for learning feedback control systems, and it has been used in many courses across multiple departments. A 3-DOF parallel robot has been used as a Distance Laboratory System \cite{santana2013remote} showing that users can collaborate on control algorithms and enhance their experience. However, these online laboratories either focus only on the remotization aspect with no tutoring content or lack step-by-step guidance for the students to learn new concepts. 

% The cloud robotics project \cite{hu2012cloud}, the RoboEarth project \cite{waibel2011roboearth}\cite{hunziker2013rapyuta} and its follow-up project from Rapyuta Robotics \cite{Rapyuta}, have developed cloud based platforms to share information and knowledge between robots. These type of platforms aim to ease non-expert users to manage and utilize state-of-the-art robots.

% In contrast, our platform focuses in these two capabilities.

% In astronomy, iTelescope \cite{iTelescope} is a shared observatory which provides remote telescope hosting services for researchers and amateurs of astronomy. In life science, there is the Emerald Cloud Lab \cite{Emerald} which enables researchers to remotely conduct life science experiments via automated hardware. 

% It connects online users to 19 telescopes and observatories located in New Mexico, Australia and Spain. 

% It is accessed through a web browser and provides up to 42 different categories of life science experiments. 

% In robotics, the most notable effort is a recent project by Georgia Institute of Technology called Robotarium \cite{pickem2016robotarium}. However, these shared laboratories do not provide any educational content except for some basic beginner training. 

Open online courses, e.g. MOOCs \cite{wulf2014massive}, \cite{severance2012teaching}, provide fundamental tutoring. In addition, researchers have built several systems to acquire specific knowledge skills. Sherlock \cite{katz1998sherlock} is used for guiding students to troubleshoot electrical equipment; ELM-ART \cite{brusilovsky1996elm} supports students to learn programming in Lisp; Andes \cite{vanlehn2005andes} provides tutoring for learning physics; and Why2-Atlas \cite{vanlehn2002architecture} guides students in writing physics essays. These online courses all provide tutoring content using questions, hints, and guidelines, which assist the learners to obtain specific competencies. However, these online courses have not been connected to any experimental hardware.

There are also other great efforts to combine educational tutoring with laboratory devices. Quanser \cite{Quanser} has been founded in response to a need among engineering educators and researchers for robust, high precision control experiments. The company developed a servomotor system featuring a modular platform for teaching and research. However, their platform is not remote as it needs to be in the same location as its users. iLabCentral \cite{iLabCentral} is an educational platform for remote hardware experimentation. The platform has been used to connect various laboratory devices for science education. However their devices provide learning of basic scientific principles instead of teaching content for state-of-the-art engineering platforms.

These challenges and opportunities have prompted the authors to develop the UT Cloud Laboratory (CLAB) with focus on teaching concepts related to state-of-the-art engineering equipment. Given the focus on distance learning, it has leveraged distributed communication software and cloud framework technologies to provide remote education with a state-of-the-art robotic actuator (Fig. \ref{fig:overview}). In Section II, the platform design, which includes a series elastic actuator (SEA) testbed and a cloud-based software architecture is introduced. It also introduces a Linear Temporal Logic (LTL) \cite{baier2008principles} method which automatically synthesize a finite state machine to manage human-server-machine interaction with a procedural control system experiment. In Section III, experiment content for system identification \cite[pg.175-178]{paine2014high} and feedback control \cite[pg.404-410]{gajic1996modern} are described. The feedback control experiment has been implemented in the CLAB system using the LTL synthesis method. In Section IV, the paper introduces a method to present to the students clean system identification plots from the experimental data by applying a piecewise Fast Fourier Transformation method. In Section V, the educational browser interface for usage of the CLAB system is shown which includes pre-laboratory questions, usage scheduling, and remote laboratory experimentation content.

\section{Platform Design}

\subsection{State-of-the-art Hardware}
\begin{figure}[t]
\centering
\includegraphics[width=.95\linewidth]{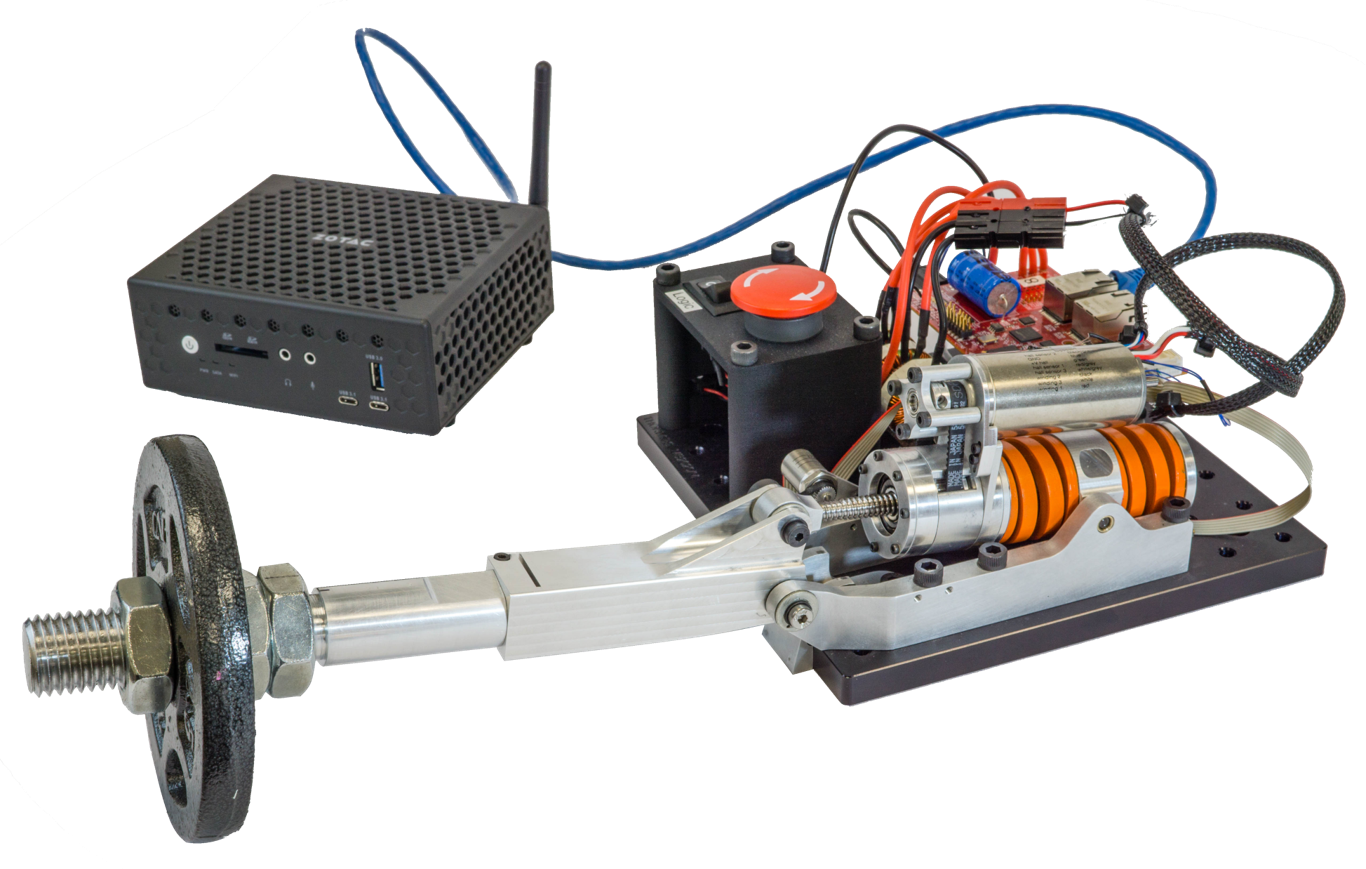}
\caption{P-170 Orion Series Elastic Actuator Testbed. The platform includes a DC brushless motor, an elastic spring, a ball screw drive transmission drive, two encoders, and a thermal sensor attached to the the surface of the motor. To control the motor and read sensor data, the testbed has an AXON control board which includes EtherCat communications, provides optimized real-time control, low latency data transfer, and a floating point processor.}
\label{fig:p170}
\end{figure}

The CLAB system includes an Apptronik P-170 Orion SEA testbed (Fig. \ref{fig:p170}). The  experiment content is suited to teach third and/or fourth year engineering students principles of control system theory. The P-170 was developed based on the UT-SEA \cite{paine2014design}, which has been adapted for the NASA Valkyrie robot \cite{radford2015valkyrie}, \cite{paine2015actuator}. The SEA's control board is connected to a small PC with Ubuntu 14.04 OS which connects to a cloud server for remotization. 

% The hardware setup is shown in Figure \ref{fig:p170}. 

\subsection{Software Architecture}

\begin{figure}[t]
\centering
\includegraphics[width=.95\linewidth]{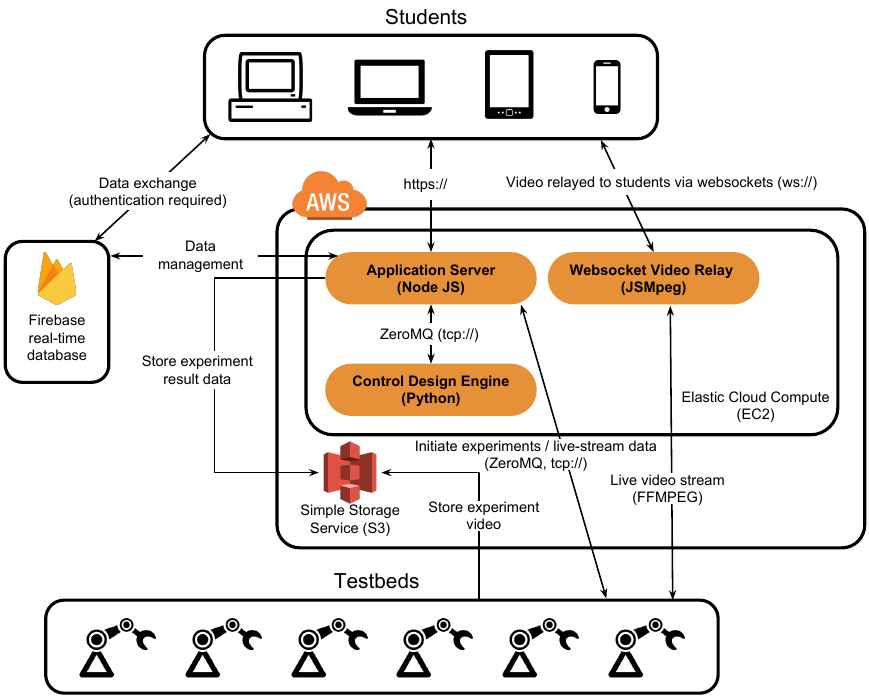}
\caption{Overview of the CLAB software system, which employs the capabilities of Google Firebase and AWS}
\label{fig:software_archi}
\end{figure}

% Do I need to ask Neal and Mara if they quote any sentences from others' word since I use their recording notes below?
% Another question is that since both of you did not mention about demo mode. So maybe I will delete the materials related with that.
Several software platforms have been established for massive open online courses\cite{breslow2013studying}, including POSA MOOC\cite{schmidt2013producing}. Fig. \ref{fig:software_archi} shows the diagram of the software system that drives the CLAB portal. The software system is supported by the Google Firebase database \cite{firebase} and Amazon Web services (AWS) \cite{AWS} in order to connect users to the P-170 SEA.

Google Firebase is a real-time database, which is a cloud-hosted and allows people to store and synchronize data between users in real-time. It offers a variety of services common to modern web applications. CLAB makes use of Firebase's authentication service and the database. In the software framework, Firebase is used for managing user accounts as well as saving data from the users. The data that is collected, including students' answers to pre-lab questions, students' scheduled experimentation times, and students' input parameters for each experiment are stored on the application.

Data to be stored in the Google Firebase real-time database is managed in the application server, which is built upon AWS. The application server acts as an intermediary between the web portal and other resources. The server communicates with the experimentation testbed to handle several activities, including reserving a block of experimentation time, initiating an experiment, and streaming experimental data for students to view. The server constructs the structure of the interface, offers different activity options to users, and changes the actuator operating modes. It also communicates with the control design engine. For instance, once a student chooses the first lab experiment, the server will transfer the mode value to align with the control design engine, and will allow the control design engine to evaluate the student's answers.

The Control Design Engine is in the form of a finite state machine synthesized by the LTL method to manage the action-reaction interactions between student, server, and testbed. For example, the range of acceptable values for an experimentation question may depend on values submitted earlier in the control design sequence. The control design engine contains the logic for dynamically determining which student submissions are acceptable. Additional details will be provided in subsequent sections.  

To improve student experience when interacting with the CLAB system, real-time video streaming is provided during the lab section. The web socket video relay takes in a live video of the lab device and broadcasts the feedback to a student in a browser-friendly format via the web portal. The video relay is optimized to be low latency. The time between a student pressing the "Start Run" button and seeing the device execution is only about two tenths of a second. JSMpeg and FFMpeg are used for the video recording feature. JSMpeg is a video player written in JavaScript. FFMpeg is a cross-platform solution to record, convert, and stream audio and video. After the experiment, the experimental data and the video streaming will be stored in Amazon's simple cloud storage service, S3, for later review.

The SEA is run by a computer with a web camera pointed at the device itself. The computer receives and responds to messages via ZeroMQ, a lightweight messaging framework, to perform actions and launch experiments. Live experimental output data is broadcast in real-time back to the application server using ZeroMQ's publisher-subscriber model. The data is subsequently forwarded to the student’s browser for real-time plotting.

\subsection{Control Design Engine}

\subsubsection{Student-Server-Machine Interaction}

In a controller design experiment, the server asks the student to enter variable values through posing lab questions. The server will check these variables within the cloud computation segment of the control design engine, and will record whether these variables have been entered correctly. When all the variables have been correctly entered and the student enters the "Start Run" command, the server tells the testbed's local computer to start the experiment, and receives the experiment results if the experiment is successfully finished. The machine reports an error message to the server if there is any problem in running the experiment. The student can reset the system back to the beginning of the controller design activity at any time.

\begin{figure}[t]
\centering
\includegraphics[width=\linewidth]{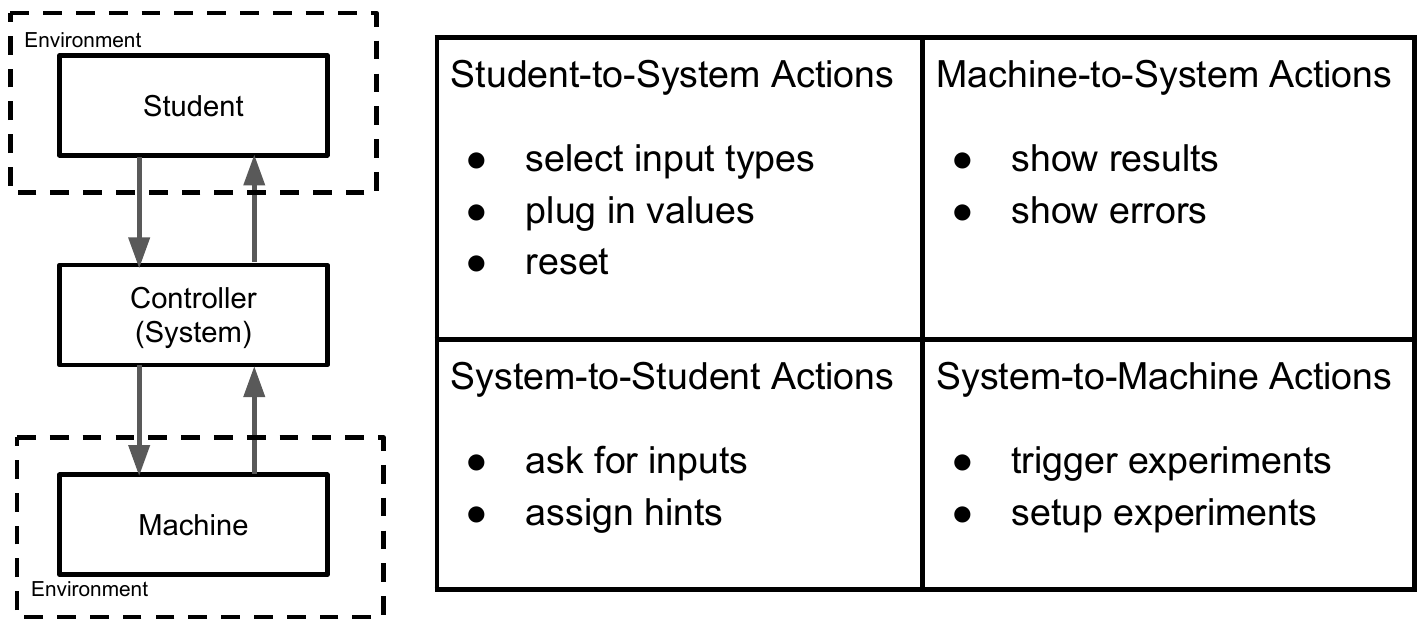}
\caption{An overview of the information communicated from one CLAB component to another, where the key components are the student at the high level, the cloud computation at the mid-level, the experiment device (machine) at the low level, and the Control Design Engine (system) serving as an intermediary.}
\label{fig:interactions}
\end{figure}

The Control Design Engine, which lives in the Amazon server, manages all interactions between the student, cloud computation, and local machine on the testbed (Fig. \ref{fig:interactions}). Whenever there is an action executed from the student or machine, the Control Design Engine identifies the correct responding action to guide the student to complete all questions and collect all necessary inputs to start the experiment. 

% Figure \ref{fig:interactions} shows the list of actions between the Control Design Engine, human, cloud computation, and machine. 

% Rachel starts to review

\subsubsection{Linear Temporal Logic}

The Control Design Engine, which is called "system" in this LTL synthesis, is a finite state machine in which transitions are triggered from the student, cloud computation, and machine actions. The state indicates the suitable system response. LTL \cite{baier2008principles} is used to represent the system specifications for all interactions, and automatically synthesizes the finite state machine.

First, the atomic propositions (AP) of LTL are defined based on the actions of the system, student, cloud computation, and machine. Assume $x$ is one variable from a series of questions that the student needs to complete before the experiment. $s$, $h$, $c$, and $m$ stand for system, student, cloud computation, and local machine. The associated atomic propositions are as follows: 

\begin{enumerate}

\item System-to-Human actions: $\varphi^{s-h}_{x}$ indicates the system asks a question to the student to calculate $x$. $\varphi^{s-h}_{exp}$ indicates the system asks if the student wants to enter the experiment. $\varphi^{s-h}_{reset}$ indicates the system asks if the student wants to reset itself. All system-to-human action APs are included in set $S_{h}$.

\item Human-to-System actions: $\varphi^{h}_{x}$ indicates the student provides the value of $x$ to the system; $\varphi^{h}_{exp}$ indicates the student clicks the "Start Run" button to run the experiment; $\varphi^{h}_{reset}$ indicates the student clicks the reset button to restart from the beginning. All human-to-system action APs are included in set $E_{h}$.

\item System-to-Computation actions: $\varphi^{s-c}_{x}$ indicates the cloud computation checks whether $x$ has been correctly calculated by the student. All system-to-computation action APs are included in set $S_{c}$.

\item Computation-to-System actions: $\varphi^{c}_{true}$ and $\varphi^{c}_{false}$ indicate the correctness of cloud computation checking result for the entered variable. All computation-to-system action APs are included in set $E_{c}$.

\item System-to-Machine actions: $\varphi^{s-m}_{exp}$ indicates that the system triggers the experiment on the low level testbed machine. All system-to-machine action APs are included in set $S_{m}$.

\item Machine-to-System actions: $\varphi^{m}_{result}$ indicates that the experiment has been completed successfully and the experimental results have been displayed and saved; $\varphi^{m}_{error}$ indicates the experiment has been interrupted and an error message has been collected. All machine-to-system action APs are included in set $E_{m}$.

\item System-Self actions: $\varphi^{s}_{h}$, $\varphi^{s}_{m}$, and $\varphi^{s}_{c}$ indicate which external factor the system is currently communicating with. $\varphi^{s}_{x}$ indicates the variable $x$ has been calculated correctly by the student and is saved for the setup of the experiment. $\varphi^{s}_{result}$ indicates experiment results have been saved in the system. 

\end{enumerate}

For ease of later discussions, superscript $e \in \{h, m, c\}$ is also defined.

% one of the external factors which plays key roles in driving the operation of the CLAB software. We also define set $I$, which includes all the actions from the external factors. \rt{(Is set $I = \{h, c, m\}$? or is it $I = \{x, exp, rst, T, F\}$?)}

All APs are formulated into constraints by propositional logic operators and temporal operators for automatic synthesis. The temporal operators, $\Box$ stands for 'always', $\Diamond$ stands for 'eventually', and $\bigcirc$ stands for 'next time step.'

All interactions are formulated under a structure of a two-player game in which one player is the system and the other player is the group of all external factors (human, cloud computation, and local machine). The system specifications of the two-player game are expressed into GR(1) \cite{piterman2006synthesis} formulas, which allows automatic synthesis of the finite state machine. The GR(1) formulas include safety assumptions, liveness assumptions, and initial assumptions for human, computation and machine; and safety requirements, liveness requirements, and initial requirements for the server. The assumptions of actions from human, cloud computation, and machine are defined as the follows.

\begin{enumerate}

\item Safety Assumption 1: At each time step, there is only one action $\varphi^{e}_{i} \in{E_h \cup E_c \cup E_m}$ executed by the student, computation, or machine, 

\begin{equation}
\Box \quad \varphi^{e}_{i} \rightarrow \neg \bigvee_{j \neq i, \, \varphi^{e}_{j} \in{E_h \cup E_c \cup E_m}} \varphi^{e}_{j} 
\end{equation}

\item Safety Assumption 2: The student
%Student 
will not be able to enter the value of $x$ in the next time step if a variable $x$ has not been requested by the server at the current time step, 

\begin{equation}
\Box \quad \neg \varphi^{s-h}_{x} \rightarrow \bigcirc \neg \varphi^{h}_{x} 
\end{equation}

\item Safety Assumption 3: The student
%Student 
will not be able to enter the experiment in the next time step if the experiment has not been requested by the server at the current time step, 

\begin{equation}
\Box \quad \neg \varphi^{s-h}_{exp} \rightarrow \bigcirc \neg \varphi^{h}_{exp} 
\end{equation}

\item Safety Assumption 4: The student
%Student 
will not be able to reset the system in the next time step if a reset has not been requested by the server at the current time step, 

\begin{equation}
\Box \quad \neg \varphi^{s-h}_{reset} \rightarrow \bigcirc \neg \varphi^{h}_{reset} 
\end{equation}

\item Safety Assumption 5: The machine
%Machine 
will send experiment results or an error message in the next time step if and only if the server has triggered the experiment at the current time step,
%Actually sometimes the machine will send both results and error back. For example, it will update the data if data is transferred correctly as well as send camera error if the camera isn't turned on.

\begin{equation}
\Box \quad \varphi^{s-m}_{exp} \leftrightarrow \bigcirc (\varphi^{m}_{result} \vee \varphi^{m}_{error})
\end{equation}

\item Safety Assumption 6: The cloud computation
% (module? factor?)}
%Computation 
will send a `correct' message or `incorrect'
%'correct' message or 'incorrect' 
message in the next time step if and only if the server has asked the computation to check the value of $x$ at the current time step,

\begin{equation}
\Box \quad \varphi^{s-c}_{x} \rightarrow \bigcirc (\varphi^{c}_{true} \vee \varphi^{c}_{false})
\end{equation}

\item Liveness Assumption 1: Any action, $\varphi^{e}_{i} \in{E_h \cup E_c \cup E_m}$, from the student, computation, and machine is available in the future at every time step,

\begin{equation}
\Box \Diamond \quad \varphi^{e}_{i}
\end{equation}

\end{enumerate}

The system requirements from the server are defined as follows, noting that there is no liveness requirement in these specifications:
% \begin{figure}[t]
% \centering
% \includegraphics[width=0.5\linewidth]{Selection_245}
% \caption{Locked Output SEA Model \cite[pg.67]{paine2014high}}
% \label{fig:sea_model}
% \end{figure}

\begin{enumerate}

\item Safety Requirement 1: $x_1$ is a prerequisite value of value $x_2$. The server will ask the student to enter the value of $x_2$ if and only if in the student mode there is no correct value of $x_2$ saved in the server and a correct value of $x_1$ saved,

\begin{equation}
\Box \quad ( \neg \varphi^{s}_{x_2} \wedge \varphi^{s}_{x_1} \wedge \varphi^{s}_{h}) \leftrightarrow \varphi^{s-h}_{x_2} 
\end{equation}

\item Safety Requirement 2: $y$ is a prerequisite value of the experiment. The server will ask the student to enter the experiment if and only if in the student mode there is a correct value of $y$ saved in the server,

\begin{equation}
\Box \quad (\varphi^{s}_{y} \wedge \varphi^{s}_{h}) \leftrightarrow \varphi^{s-h}_{exp} 
\end{equation}

\item Safety Requirement 3: The server will ask the student to reset if and only if the server is in the student mode,

\begin{equation}
\Box \quad \varphi^{s}_{h} \leftrightarrow \varphi^{s-h}_{reset} 
\end{equation}

\item Safety Requirement 4: The server will ask the computation to check the value of a variable $x$ if and only if the student has inputted the value of $x$ at the same time step,

\begin{equation}
\Box \quad \varphi^{h}_{x}  \leftrightarrow \varphi^{s-c}_{x}
\end{equation}

\item Safety Requirement 5: The server will trigger the experiment if and only if the student has clicked the `start' button to start the experiment,

\begin{equation}
\Box \quad \varphi^{h}_{exp} \leftrightarrow \varphi^{s-m}_{exp} 
\end{equation}

\item Safety Requirement 6: The system is in the mode of machine if and only if the system formulates a machine request at the current time step. The system is in the mode of computation if and only if the system formulates a computation request at the current time step. The system is in the mode of human if and only if the system is not in the mode of machine or computation.

\begin{equation}
\begin{aligned}
& \Box \quad \varphi^{s}_{m} \leftrightarrow \bigvee_{\varphi^{s-m}_{i} \in{S_m}} \varphi^{s-m}_{i}, \quad & \Box \quad \varphi^{s}_{c} \leftrightarrow \bigvee_{\varphi^{s-c}_{i} \in{S_c}} \varphi^{s-c}_{i}, \\
% & \Box \quad \varphi^{s}_{c} \leftrightarrow \bigvee_{i \in{C}} \varphi^{c}_{i} \\
& \Box \quad (\neg \varphi^{s}_{m} \wedge \neg \varphi^{s}_{c}) \leftrightarrow \varphi^{s}_{h} 
\end{aligned}
\end{equation}

\item Safety Requirement 7: The system will hold a value of $x$ in the next time step if and only if all the follow conditions hold:

\begin{enumerate}[i.]
\item An $x$ is held by the system at the current time step or an $x$ will be entered by the student in the next time step.
\item No $x$ under the checking of cloud computation at the current time step or no `false' message will be sent from cloud computation in the next time step.
\item No `error' message will be sent from the machine in the next time step.
\item No reset command will be entered from the student in the next time step.
\end{enumerate}

\begin{equation}
\begin{aligned}
% \resizebox{.9\hsize}{!}
% {$
\Box \quad & (\varphi^{s}_{x} \vee \bigcirc \varphi^{h}_{x} ) \wedge ( \neg \varphi^{s-c}_{x} \vee \neg \bigcirc \varphi^{c}_{false} ) \\
& \wedge \neg \bigcirc \varphi^{m}_{error} \wedge \neg \bigcirc \varphi^{h}_{reset} \leftrightarrow \bigcirc \varphi^{s}_{x}
% $}
\end{aligned}
\end{equation}

\item Safety Requirement 8: The system will hold an experiment result in the next time step if and only if all the following conditions hold:

\begin{enumerate}[i.]
\item An experiment result is held by the system at the current time step or an experiment result will be sent from the machine in the next time step.
\item No reset command will be entered by the student in the next time step.
\end{enumerate}

\begin{equation}
\resizebox{.8\hsize}{!}
{$
\begin{aligned}
& \Box \quad (\varphi^{s}_{result} \vee \bigcirc \varphi^{m}_{result} ) \wedge \neg \bigcirc \varphi^{h}_{reset} \leftrightarrow \bigcirc \varphi^{s}_{result} \\
\end{aligned}
$}
\end{equation}

% \rt{(Maybe you don't have enough room anymore, but the safety requirement 5 and 6 are lacking description for the temporal logic statements.) }

\end{enumerate}

% \begin{figure}[t]
% \centering
% \includegraphics[width=1.\linewidth]{exp2-2.pdf}
% \caption{Topology of a simplified automaton (7 environment actions, 10 system actions, and 26 states) for an experiment. The arrows determine system actions which trigger the states they point to.}
% \label{automata}
% \end{figure}

% \begin{figure*}[t]
% \centering
% %\includegraphics[width=.9\linewidth]{example_problem_1.pdf}
% \includegraphics[width=.9\linewidth]{problem.PNG}
% % * <lsentis@gmail.com> 2018-01-07T22:56:47.246Z:
% %
% % ^.
% \caption{(a) An example system identification prelab question. The purpose of this question is to ensure that the student can model dynamic SEA system and express the system's input-output relationship in the Laplace domain. (b) An example feedback control prelab question. This question drives the student to ensure his or her ability to interpret Bode diagram results and to understand the meaning of the controller design parameter $\phi_{max}$.} 
% \label{fig:prelab}
% \end{figure*}

Finally, after defining all these assumptions and system requirements, GR(1) is used to solve a finite state machine which satisfies all constraints. For the CLAB system, TuLip \cite{wongpiromsarn2011tulip}, which is a Python LTL synthesis package, was used to formulate and solve all the constraints discussed. The resulting finite state machine 
% (Fig. \ref{automata})
was implemented as the Control Design Engine in the server.

Notice that initial assumptions and requirements have not been specified. However, in order to ensure the system starts properly, reset commands are automatically executed at the beginning of the lab questions.

% Rachel ends review

\section{Example Experiments}

In this section, two example experiments for teaching third and/or fourth year engineering students control systems theory has been provided. The first experiment, system identification, is a preliminary task before all other experiments on the testbed. The lab questions designed for the second experiment, feedback control, provide the students step-by-step guidance to design a lead controller. The automatic synthesis of the finite state machine under the constraints (1) - (11) will be applied to ease the implementation of the feedback control experiment on CLAB. The sequence of the lab questions are programmed under the form of Safety Requirement 1 in the previous section.

\subsection{System Identification}

Before students design control systems for the actuator, they need to understand the system model and its mathematical representation. When the actuator arm has been locked, it is modeled as a second order linear system (Fig. \ref{fig:prelab}) \cite[pg.75]{paine2014high} with a transfer function:

\begin{equation}
P(s) = \frac{\beta k_{s}}{m_k s ^ 2 + b_{eff} s + k_{s}},
\end{equation}

where $m_k$ is the spring-motor lumped mass, $b_{eff}$ is the effective damping ratio of the spring, $k_{s}$ is the spring constant, and $\beta$ is a product of the motor efficiency and gear ratio of the drivetrain. 

In the system identification experiment \cite[pg.175-178]{paine2014high}, electric current in the form of sine waves of varying frequencies are sent to the series elastic actuator. The resulting system outputs, spring forces, are measured. From this data, a Bode diagram can be developed, showing the magnitude and phase diagrams when considering output spring force as a result of input current of varying frequencies. The Bode diagrams allow the student to measure the associated phase margin as well as to recognize the effects of non-ideal conditions within an engineering system.

% By applying constraints (1) - (11), a finite state automata with 15 environment actions, 33 system actions, and 581 states are automatically synthesized.

% \begin{figure}[ht]
% \centering
% \includegraphics[width=.6\linewidth]{Selection_182}
% \caption{System identification results with real data (orange dots) and fitting bode curve (blue)}
% \label{fig:sys_id_plot}
% \end{figure}

\subsection{Feedback Control}

\begin{figure}[t]
\centering
\includegraphics[width=0.9\linewidth]{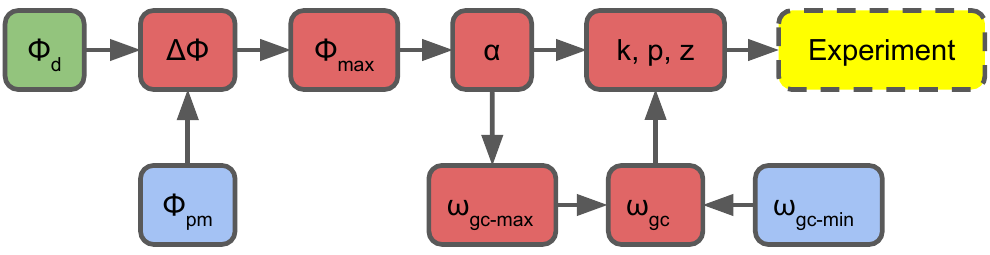}
\caption{Flow chart of the feedback control experiment. The original phase margin $\phi_{pm}$ and the original gain-crossover frequency $\omega_{gc-min}$ (both blue) are measured from the original Bode diagrams from the system identification experiment. The desired phase margin (green) is given. The other intermediate parameters (red) are asked in a step-by step manner. These questions assist the student in developing a transfer function for the lead controller such that the student can implement the lead controller to obtain a new Bode diagram to see the phase margin improvement.}
\label{fig:feedback_chart}
\end{figure}

Without a controller, this SEA will be unstable in the closed loop configuration due to the negative phase margin that was measured in the system identification experiment. The phase margin should be positive and relatively large so that it can stably track a desired output trajectory and overcome disturbances and time delay from the real world. The feedback control experiment introduces a lead controller, which is a type of feedback controller used to improve the phase margin and system stability. A series of six questions based on a control system textbook \cite[pg.404-410]{gajic1996modern} guides the student step-by-step to design the lead controller (Fig. \ref{fig:feedback_chart}). 

%Note that these steps are originally in a recursive algorithm. We only let the %students to run the algorithm with one iteration to prove the concept in the %experiment.

Before the feedback control design experiment, the Bode diagram from the system identification experiment is amplified by a predefined steady gain $k_{ss}$. The original phase margin $\phi_{pm}$, and the original gain-crossover frequency $\omega_{gc-min}$ are measured by the students from the amplified Bode diagram. The first question asks about $\Delta \phi$, which is the difference between $\phi_d$ and $\phi_{pm}$. 

\begin{equation}
\Delta \phi = \phi_{d} - \phi_{pm}
\end{equation}

The second question asks about $\phi_{max}$, which is chosen based on the entered value of $\Delta \phi$. As a design rule of thumb \cite[pg.405]{gajic1996modern}, $\phi_{max}$ is $5 - 10 ^{\circ}$ greater than $\Delta \phi$. The third question asks about $\alpha$, which is the ratio between the pole and the zero of the transfer function of the lead controller. The parameter can be calculated from a function in terms of $\phi_{max}$.

\begin{equation}
\alpha = \frac{1 + \sin{\phi_{max}}}{1 - \sin{\phi_{max}}}
\end{equation}

The CLAB system then requests that the student enter $\omega_{gc-max}$, the maximum possible value of the new gain-crossover frequency. The value of $\omega_{gc-max}$ is measured on the original Bode diagrams by finding the new gain-crossover frequency after shifting the magnitude diagram $20 \log \alpha$ upward. The fifth question asks the student about the new gain-crossover frequency $\omega_{gc}$, which is a value between $\omega_{gc-min}$ and $\omega_{gc-max}$. The final question requests that the gain $k$, pole $p$, and zero $z$ values of the lead controller transfer function be entered into the interface. The prerequisite values have all been calculated in the previous questions. 

\begin{equation}
\begin{aligned}
k = \alpha, \quad p = \omega_{gc} \sqrt{\alpha}, \quad z = \frac{\omega_{gc}}{\sqrt{\alpha}}. 
\end{aligned}
\end{equation}

The transfer function of the lead controller $G_c (s)$ with the predefined steady gain can be expressed as:

\begin{equation}
G_c (s) = k_{ss} \cdot k \cdot \frac{s + z}{s + p}
\end{equation}

Similar to the system identification experiment, electrical current in the form of sine waves of varying frequencies are the inputs and the output spring forces, are measured. The lead controller with the transfer function of (17) is automatically implemented into the experiment. The Bode diagrams are created from the input command. The current command is inversely calculated from the relationship between electrical current and output spring force provided by the controller equation. The student will be able to see that the phase margin on the bode diagram has been improved by the lead controller and be able to compare this new phase margin with the desired phase margin.

% \begin{figure}[t]
% \centering
% \includegraphics[width=.8\linewidth]{Selection_292.pdf}
% \caption{(a) A subsection of the autamata for feedback control experiment. The arrows show the label of actions which trigger the state they point to.}
% \label{automata}
% \end{figure}

By applying constraints (1) - (11), a finite state automata with 13 environment actions, 28 system actions, and 78 states is automatically synthesized.

\section{Results}
\urlstyle{tt}
The vedio of demonstration is available at \url{https://www.youtube.com/watch?v=IkcvnRCCqoI}.
Before the user enters into an experiment, the CLAB home page provides an introduction to the testbed, the CLAB interface, and the motivation for the experiments. The user interface is separated into two parts: a menu bar and the main content of each experiment. The menu bar on the left side of the CLAB interface shows the session names for each experiment, which is composed of three subsection: pre-lab, scheduler, and experiment. The pre-lab tab leads the users to understand fundamental knowledge used in the experiments with a sequence of questions. After users complete the pre-lab, they can schedule the lab time by clicking "Schedule Lab," shown in the menu bar. 

The lab activity will become unblocked in the menu bar only when the scheduled time begins, so that multiple users (with different login ID's) will not be able to access the experiment system to run the testbed at the same time. It is important to note that multiple students can access the experiment via multiple laptops and/or iPhones using the same login ID, so that use of CLAB in a classroom environment is supported. When users start the lab activity, there may be a set of questions instructing them to input the parameters  
used in running the experiment. The interface in which the experiment is executed contains real-time video streaming, real-time input and output data, and simulation results for comparison. After the experiment is complete, Bode diagrams will also be provided for comparison against the simulated results.  

After the experiment, all the data and video will be automatically stored on the S3 Amazon Server for users to conveniently review their experiments and data to make observations and conclusion about the results.

\begin{figure}[t]
\centering
%	\subfigure[The schedule interface before the user schedules an experimental time, with unavailable times indicated in gray.]{\label{schedule1}\includegraphics[width=1.\linewidth]{Selection_289}}
	\subfigure
    {\label{schedule2}\includegraphics[width=1.0\linewidth]{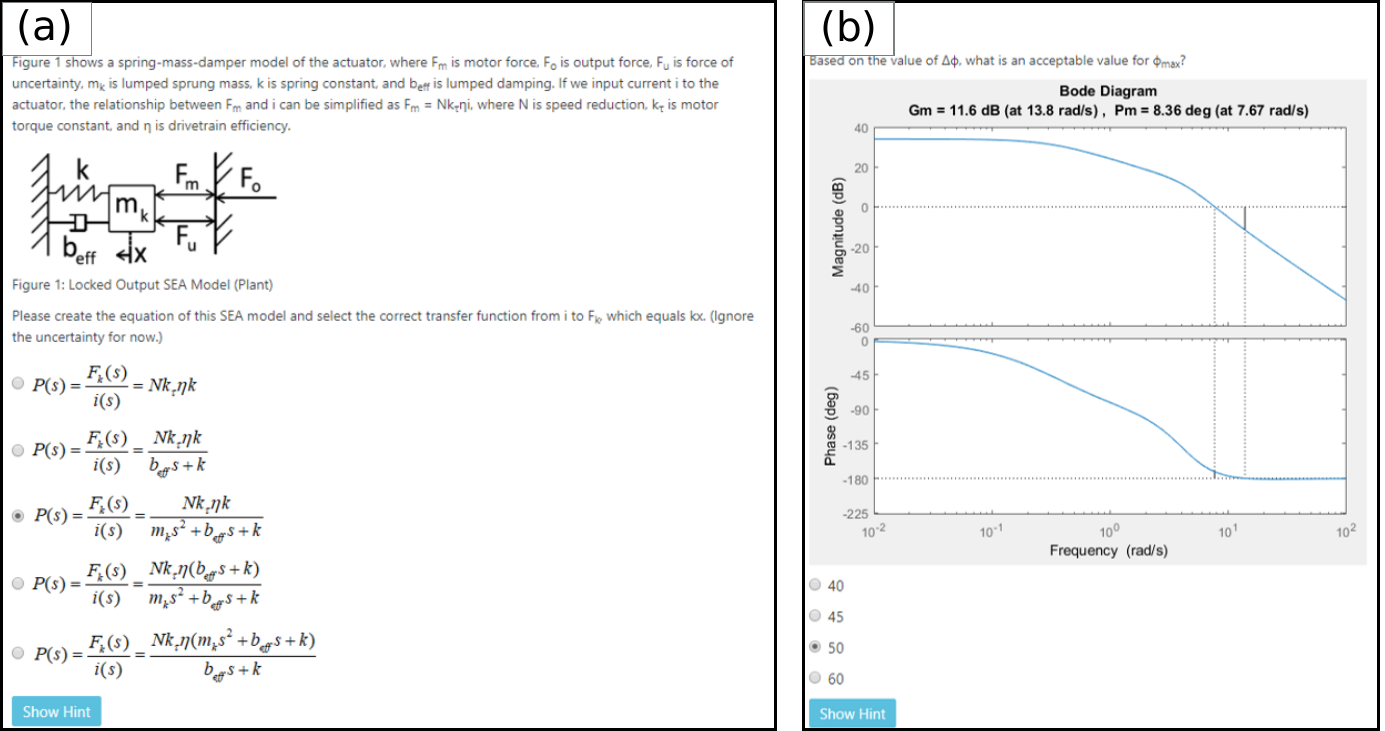}}
    %Rachel could you help me check this again
\caption{(a) An example system identification prelab question. The purpose of this question is to ensure that the student can model dynamic SEA systems and express the system's input-output relationship in the Laplace domain. (b) An example feedback control prelab question. This question drives the student to ensure his or her ability to interpret Bode diagram results and to understand the meaning of the controller design parameter $\phi_{max}$.} 
\label{fig:prelab}
\end{figure}

\begin{figure}[t]
\centering
%	\subfigure[The schedule interface before the user schedules an experimental time, with unavailable times indicated in gray.]{\label{schedule1}\includegraphics[width=1.\linewidth]{Selection_289}}
	\subfigure    {\label{schedule2}\includegraphics[width=1.0\linewidth]{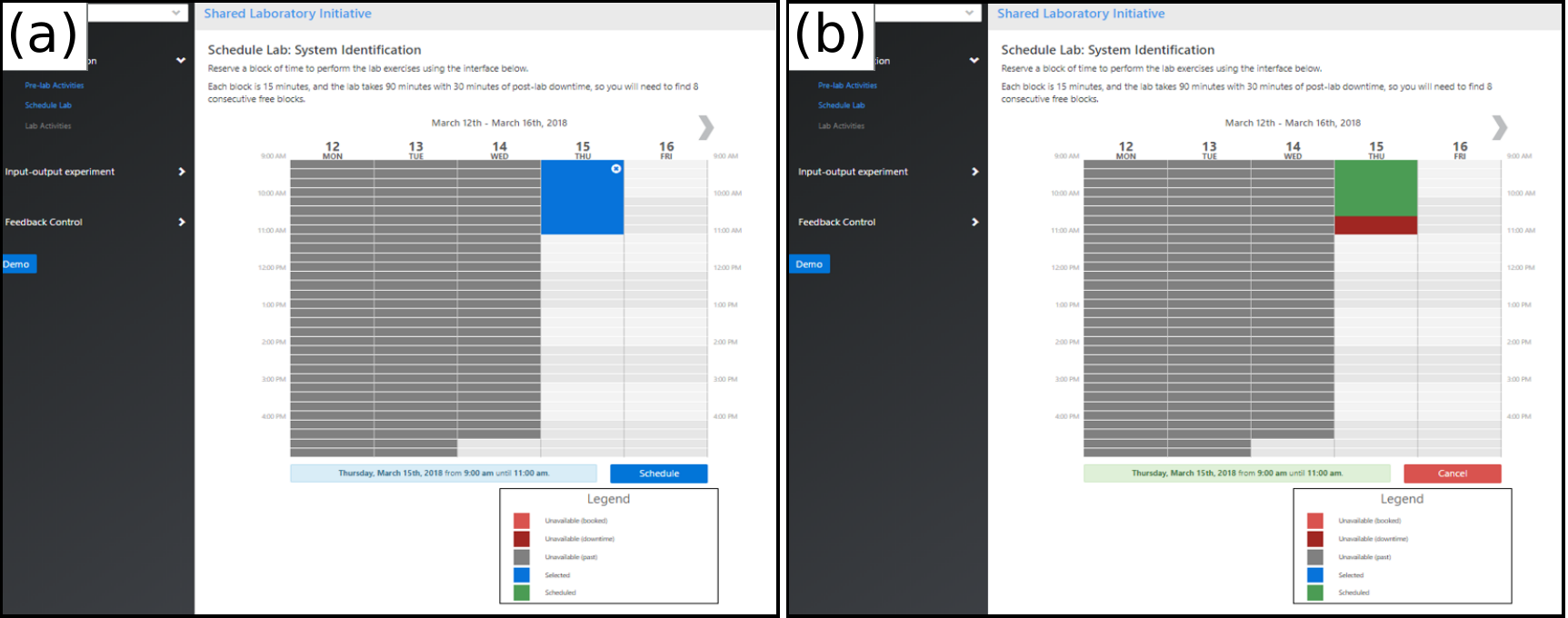}}
    \subfigure    
    {\label{schedule3}}
    %Rachel could you help me check this again
\caption{An overview of the process a user follows to reserve an experiment time. (a) The blue block indicates to the user that he or she can reserve the selected time, with unavailable times indicated in gray. (b) Once the user selects a time, the interface reserves a block suitable for experimentation (green) and hardware cool-down time (red).}
\label{scheduler}
\end{figure}

\subsection{Prelab}
The CLAB interface allows for a prelab to be associated with each experiment. Instructors and teaching assistants can create and modify the questions using Firebase. Firebase allows for the questions to be structured in three different ways: free response questions that take a specific answer, free response questions which can take a range of answers, and multiple choice questions. The purpose of the prelab is that which is typical for in-person laboratories: to ensure the understanding of key concepts which are relevant  to the current experiment so that the student can appreciate the applicability of theory to the real world. 

Example questions for the system identification and feedback control experiments are provided in Fig. \ref{fig:prelab}. In Fig. \ref{fig:prelab}(a), the locked output SEA model, which was discussed previously, is presented, and the student is asked to devise the transfer function between input current and the force measured in the spring. This question allows for the understanding of model development, transfer functions, and the input-output relationship relevant for system identification. If the student is unsure of an answer, he or she can access a hint \cite{franklin1994feedback}, \cite{keesman2011system}, which provides further insight on the relevant topics and/or a pointer to useful references for more information. Other topics covered in the system identification prelab relate to experiment procedure, system behavior as a function of different degrees of damping, system order, and phase margin measurement. If the answer is submitted incorrectly, the student can continue to submit answers until it is correct. 

Fig. \ref{fig:prelab}(b) shows the interface for the feedback control prelab. The prelab focuses on ensuring that the student understands the relevance of lead controller design parameters and the ability to formulate these parameters, as demonstrated in the question displayed. Users can move forward as long as their responses are in the range of accepted answers, and they are allowed to go back and review their answers after they have answered correctly. The ability to schedule a lab is contingent upon full completion of the associated prelab. Upon completion, the student is automatically directed to the scheduler. 

% \begin{figure}[t]
% \centering
% %	\subfigure[The schedule interface before the user schedules an experimental time, with unavailable times indicated in gray.]{\label{schedule1}\includegraphics[width=1.\linewidth]{Selection_289}}
% 	\subfigure[The blue block indicates to the user that he or she can reserve the selected time, with unavailable times indicated in gray.]{\label{schedule2}\includegraphics[width=.8\linewidth]{Selection_290}}
%     %Rachel could you help me check this again
%     \subfigure[Once the user selects a time, the interface reserves a block suitable for experimentation (green) and hardware cool-down time (red).]{\label{schedule3}\includegraphics[width=.8\linewidth]{Selection_291}}
% \caption{An overview of the process a user follows to reserve an experimental time.}
% \label{scheduler}
% \end{figure}

\subsection{Scheduler}

The CLAB scheduler interface guides users to schedule the experiments online. The scheduler helps multiple users efficiently make use of the device. The scheduler interface design does not change between experiments, and the feedback control experiment is provided as an example.

Fig. \ref{scheduler} shows the interface %before, 
during and after scheduling an experiment time. Fig. \ref{schedule2} shows that times occupied by others or which have already passed are denoted as gray, and these periods are automatically blocked for scheduling. The user can click on the unblocked spaces to select a time between 9:00 AM to 4:30 PM. A legend box is provided to distinguish between blocked times, times for ``actuator cool-down time" in order to not stress the machine, and time reserved by the user. After clicking on the calendar, the system will automatically reserve two hours for the experiment section.%, as shown in Fig. \ref{schedule2}. %Rachel could you help me check this again
The period selected will be highlighted in blue. If users are satisfied with the time period they can click on the ``Schedule" button located in the bottom-right of the interface. After a user schedules the experiment time, that period will turn green to indicate that the time has been reserved, as shown in Fig. \ref{schedule3}. After the reservation is complete, the ``Schedule" button changes into a ``Cancel" button to allow students to change their reserved times, if needed. Users are allowed to cancel their appointment until the beginning of the lab section. By clicking on the ``Cancel" button, the previously-reserved time is unreserved, and the user is allowed to choose another other experiment time period. 

There are multiple rules to conveniently help users manage their scheduling: Users cannot arrange another time period for the same experiment until they finish the one they already scheduled; Users are allowed to use multiple devices to access to the experiment website; Users are allowed to arrange multiple lab experiments if they finish all the corresponding prelab sections; Users cannot cancel the scheduling during the lab experiment section. Once the scheduling time begins, users are allowed to start their lab experiment and they can start or end at any time during the scheduling period.

% Update: ADD: Prelab 
% Update: ADD: Software Architecture Schematic
% Update: ADD Scheduler discussion  
% UPdate: ADD Finite State Machine/Linear Temporal logic (execution in Python) 

% Update: other details to discuss
% prelab details
% differences between ideal and real experimental results 
% results 
% discussion
% future work
% things that can be observed during system idetification. sounds that can be heard

\begin{figure}[t]
\centering
\includegraphics[width=1.0\linewidth]{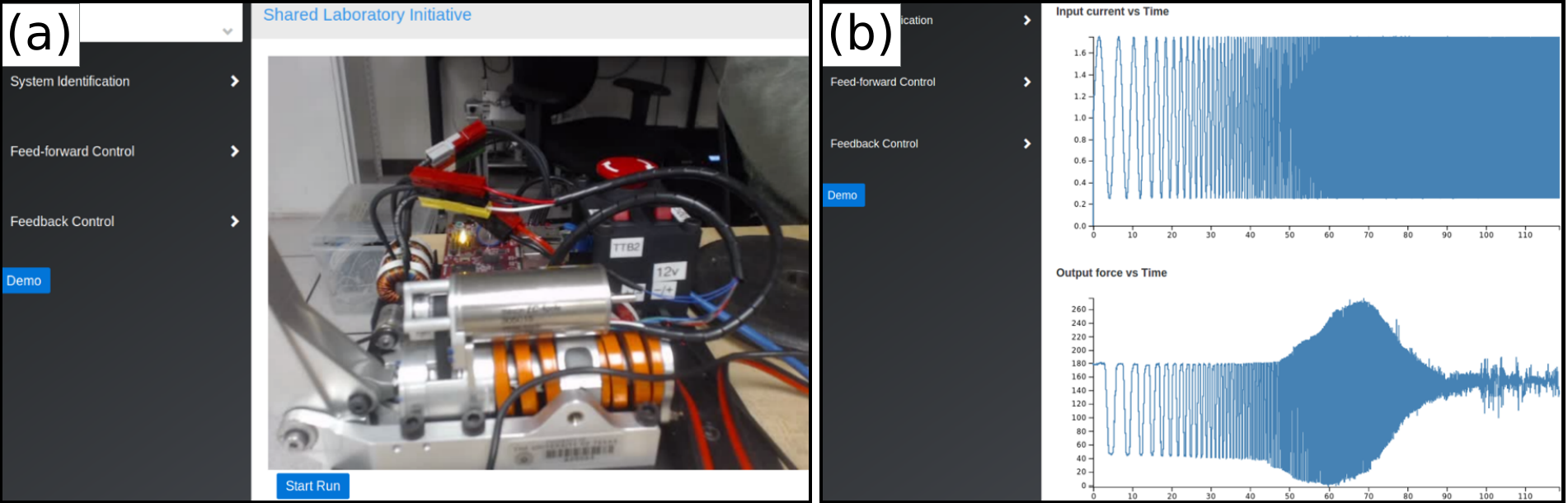}
\caption{(a) Video streaming of the P170 SEA on the CLAB interface. During the experiments, the student will notice the drive belt (connecting motor to output) moving at different frequencies, and the motions of the other components. (b) Time-domain output from the system identification experiment. The student will notice that the input current's amplitude remains constant but that the frequency increases over time. In the Output Force vs. Time plot, the varying spring force as a result of the varying frequencies provides valuable data for the generation of the associated Bode diagrams.}
\label{live_data}
\end{figure}

 \begin{figure}[t]
\centering
\includegraphics[width=1.0\linewidth]{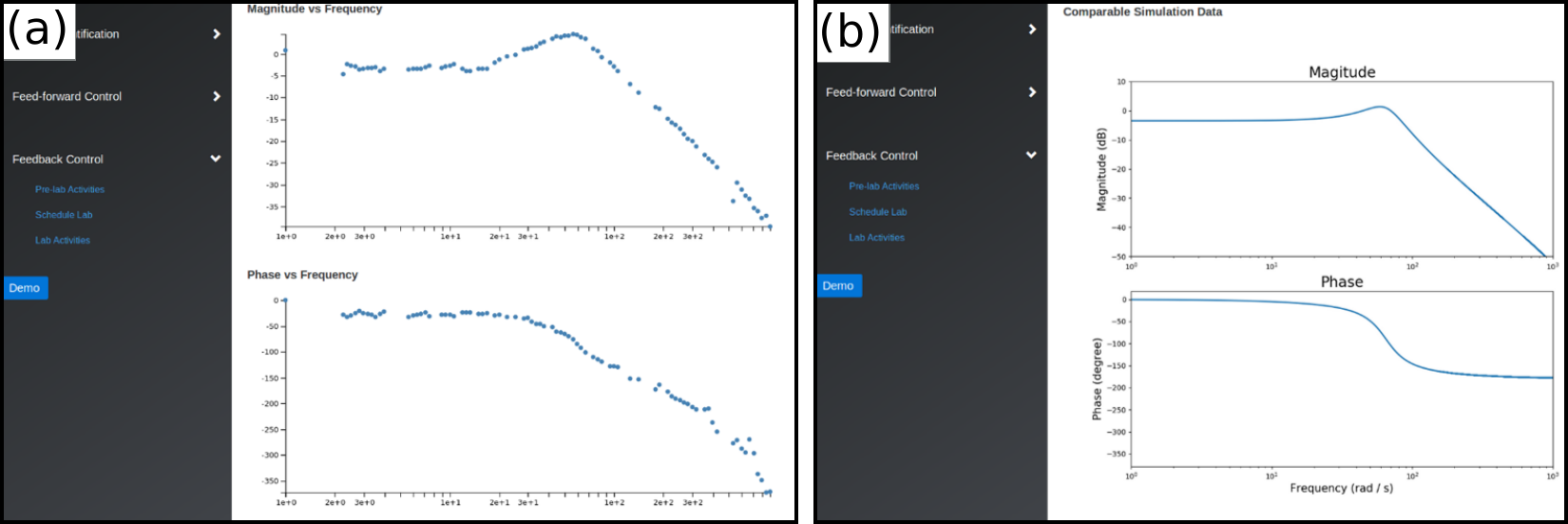}
\caption{Within the CLAB interface the student has access to plotted real data for the feedback control experiment, as well as simulation data for comparison. The student can read the positive phase margin gained from using feedback control, as well as identify deviations from expected behavior. Through side-by side comparison, one notes deviations from a predicted phase of zeros degrees at low frequencies, and deviations from a relatively constant phase at high frequencies in ideal conditions.}
\label{realVSideal}
\end{figure}

\subsection{Lab}

The CLAB interface is configured for remote access to the system identification and feedback controller design experiments. When the student conducts an experiment, he or she can see the experiment being executed in real-time (Fig. \ref{live_data}(a)). There is also audio available, as the sound of the system combined with visual information can give insight into understanding expected versus actual behaviors. For example, the student might be able to identify that the system needs to be lubricated if the belt does not move significantly despite high-frequency sounds. As the experiment is executed, time-domain data is simultaneously plotted, showing input current versus time and measured spring force versus time (Fig. \ref{live_data}(b)).

After the experiment concludes, the CLAB software automatically conducts a Fast-Fourier transform on the experimental data to produce Bode diagrams. The Bode phase diagram associated with the system identification experiment shows a negative phase margin, indicating instability. The student can then understand the benefit of implementing control of the system. The Bode diagrams of the open-loop system with the lead controller integrated into the control scheme, as seen in Fig. \ref{realVSideal}, shows the positive phase margin achieved.

%\begin{figure}[t]
%\centering
% \includegraphics[width=0.9\linewidth]{sys_id_exp_bode_plot.pdf}
% \caption{Example output from the system identification experiment. The student is able to read that the phase margin is negative and conclude that the open-loop system is unstable.}
%\label{Sys_id_bode}
% \end{figure}
 
%  \begin{figure}[t]
% \centering
%  \includegraphics[width=0.9\linewidth]{sysid.PNG}
%  \caption{Example output from the system identification experiment. The student is able to read that the phase margin is negative and conclude that the open-loop system is unstable.}
% \label{Sys_id_bode}
%  \end{figure}

In both experiments, the student has access to predicted ideal Bode diagrams which were produced in Matlab simulations. These figures for comparison highlight one of the key benefits of performing laboratory experiments on real hardware - the ability to identify the disparities between predicted and real-world behaviors. For example, as can be seen in Fig. \ref{realVSideal} for the feedback control experiment, simulation suggests that in ideal conditions the phase should be equal to zero at low frequencies. In the real system, one recognizes that the phase is less than zero at low frequencies. Disturbances to the system, such as nonlinear drivetrain friction, contribute to this observed decrease in stability. The simulation also shows that at high frequencies the ideal phase will reach a steady value as frequency increases, while the real data shows the phase continuing to decrease. Another source of imperfection in the embedded system is time delay, which may contribute to this deviation from expected behaviors. The two experiments allow for the development of key engineering skills - to decipher between predicted and actual outcomes, and to recognize the potential sources for imperfections in physical systems.   

\section{Conclusions}

In this paper, a method to learn the usage of stat-of-the-art robotics equipment via remote browsers has been discussed. The concept by devising a cloud-based infrastructure, CLAB, which delivers the experiment content pertaining to the analysis and control of a series elastic actuator developed for NASA to remote users has been demonstrated. The CLAB infrastructure allows users to answer prelab questions, obtain hints if needed, schedule lab time, solve lab questions, process real-time remote experiments, and collect experimental data and video through a web browser. With this foundation, system identification and feedback control experiments were implemented on the CLAB infrastructure. The authors believe this infrastructure could be effective at training students for their job careers since it uses state-of-the-art equipment.

% In the future, a wide variety of tutoring material for  experimental learning and include new devices into the CLAB system will be devised. Analysis of student learning response to test the benchmark the infrastructure against classical laboratory education where students learn in physical contact with the equipment will also be performed. It will focus on deploying the web application for usage into various courses in higher-ed facilities across the country and in secondary education centers. Intelligent tutoring systems involving online data analysis and cognitive decision methods can also be considered to improve the deliver of the educational content.

\section*{Acknowledgements}
This work was supported by the Longhorn Innovation Fund for Technology (LIFT) grant and the Academic Development Funds from the University of Texas at Austin, and NASA Space Technology Research Fellowship 80NSSC17K0188.
% We would also like to thank our colleagues from the Human Centered Robotics Lab and the Faculty Technology Studio, University of Texas at Austin who provided insight and expertise that assisted the research.

\bibliographystyle{IEEEtran}
\bibliography{sample}

% \section*{Author contributions statement}

% Must include all authors, identified by initials, for example:
% A.A. conceived the experiment(s),  A.A. and B.A. conducted the experiment(s), C.A. and D.A. analysed the results.  All authors reviewed the manuscript. 

% \section*{Additional information}

% To include, in this order: \textbf{Accession codes} (where applicable); \textbf{Competing financial interests} (mandatory statement). 

% The corresponding author is responsible for submitting a \href{http://www.nature.com/srep/policies/index.html#competing}{competing financial interests statement} on behalf of all authors of the paper. This statement must be included in the submitted article file.

% \begin{figure}[ht]
% \centering
% \includegraphics[width=\linewidth]{stream}
% \caption{Legend (350 words max). Example legend text.}
% \label{fig:stream}
% \end{figure}

% \begin{table}[ht]
% \centering
% \begin{tabular}{|l|l|l|}
% \hline
% Condition & n & p \\
% \hline
% A & 5 & 0.1 \\
% \hline
% B & 10 & 0.01 \\
% \hline
% \end{tabular}
% \caption{\label{tab:example}Legend (350 words max). Example legend text.}
% \end{table}

% Figures and tables can be referenced in LaTeX using the ref command, e.g. Figure \ref{fig:stream} and Table \ref{tab:example}.

\end{document}